\documentclass[sigconf]{acmart}
\AtBeginDocument{%
  }
\setcopyright{none}

\usepackage{amsmath,amsfonts,graphicx,hyperref,booktabs,xcolor}

\usepackage{amssymb}

\usepackage{caption}
\usepackage{subcaption}
\usepackage{enumitem}

\acmConference{}{}{}
\acmDOI{}

\author{
  Samin Khan \\
  Stanford Graduate School of Education \\ \texttt{samin.khan995@gmail.com}
}

\begin{document}

\title{Using Vision + Language Models to Predict Item Difficulty}

\begin{abstract}
This project investigates the capabilities of large language models (LLM) to determine the difficulty of data visualization literacy test items \citep{boyle2020}. We explore whether features derived from item text (question and answer options), the visualization image, or a combination of both, can predict item difficulty (proportion of correct responses) for U.S. adults. We use GPT-4.1-nano \citep{openai2023}, to analyze items and generate predictions based on these distinct feature sets. The multimodal approach, using both visual and text features, yields the lowest mean absolute error (MAE) (0.224), outperforming the unimodal vision-only (0.282) and text-only (0.338) approaches. The best-performing multimodal model was applied to a held-out test set for external evaluation and achieved a mean squared error of 0.10805, demonstrating the potential of LLMs for psychometric analysis \citep{embretson2013} and automated item development \citep{irvine2017}.
\end{abstract}


\maketitle

\section{Introduction}
Data visualization literacy (DVL) is an increasingly vital skill in today's information-rich society, enabling individuals to effectively interpret and communicate data presented visually \citep{ware2012}. Assessing DVL is critical for educational purposes, but developing standardized, reliable, and well-calibrated test items is challenging \citep{haladyna2013}. A key aspect of psychometric evaluation is predicting item difficulty, defined as the proportion of test-takers answering correctly \citep{lord1980}.

This project uses a dataset comprising responses from U.S. adults and college students to items from five different DVL assessments (WAN, GGR, BRBF, VLAT, CALVI) \citep{galesic2011, boy2014, lee2017, ge2023}. The goal of this project is to build predictive models for item difficulty based on characteristics of the items themselves, specifically focusing on the information available from the data visualization image, the question text, and their combination. We formulate two primary research questions:
\begin{enumerate}[noitemsep, nolistsep]
  \item What predicts data visualization literacy item difficulty most effectively: visual features (data viz image) or textual features (question text and possible responses)?
  \item What is the predictive power of combining both visual and textual features?
\end{enumerate}
By addressing these questions, we aim to gain insights into the cognitive demands of DVL tasks and explore the feasibility of using modern tools, such as multimodal LLMs, to automate or assist in psychometric analysis and test item development.

\section{Methods}
\subsection{Data Preparation}
We use the data visualization item response dataset collected by Verma and Fan (2025) \cite{verma2025difficulty}. The dataset contains participant responses to test items, including $item\_id$, $image\_url$, $question\_text$, and $possible\_responses$. Each item was associated with multiple participant responses. To obtain item-level difficulty, we aggregated the data by $item\_id$ and calculated the mean of $incorrect\_response$ (where 1 indicates an incorrect response). This metric represents the item's difficulty score, ranging from 0 (everyone correct) to 1 (everyone incorrect). We then defined $easiness$ as 1 - $difficulty$, representing the proportion of participants who answered correctly. The initial dataset was split into 80\% for validation and a 20\% held out test set. Our modeling and evaluation steps were conducted on a subset (N=154) of the total validation set (N=184) that was filtered for images that were PNGs for convenience of image processing. The test set was reserved for a final performance assessment.

\subsection{Modeling Approach}
We used the GPT-4.1-nano model via the OpenAI API \citep{openai2023}, leveraging its multimodal capabilities. We defined Pydantic models to structure the desired JSON output, ensuring reliable data extraction from the LLM responses. Three distinct modeling approaches were devised:
\begin{enumerate}[noitemsep, nolistsep]
  \item \textbf{Text-only Model:} This model uses only the $question\_text$ and $possible\_responses$ as input to the LLM. The system prompt instructs the LLM to analyze textual features like cognitive task type, question clarity, information integration level, number of options, correct answer text, distractor plausibility, and format consistency, and provide an estimated $prediction$ (proportion correct).
  \item \textbf{Vision-only Model:} This model uses only the $image\_url$ as input. The system prompt instructs the LLM to analyze visual features like chart type, axis clarity, data encoding clarity, readability, clutter level, data series count, annotations, and overall visual complexity, and provide an estimated $prediction$.
  \item \textbf{Multimodal (Vision + Text) Model:} This model uses both the $image\_url$, $question\_text$, and $possible\_responses$ as input. The system prompt instructs the LLM to perform a comprehensive analysis, considering visual elements, textual demands, answer option quality, and their interaction, to provide a estimated $prediction$.
\end{enumerate}
\subsection{Performance Evaluation}
To compare the performance of the three models, we calculated the MAE between the predicted easiness score and the actual easiness score (proportion correct) for each item in the evaluation subset. MAE is calculated as the average of the absolute differences between predicted and actual values. 
\noindent Code is available at: \url{https://github.com/samin-khan/multimodal_item_difficulty_prediction_notebook/tree/main}

\section{Results}
\subsection{Model Performance}
\begin{figure}[h]
  \centering
  \includegraphics[width=\linewidth]{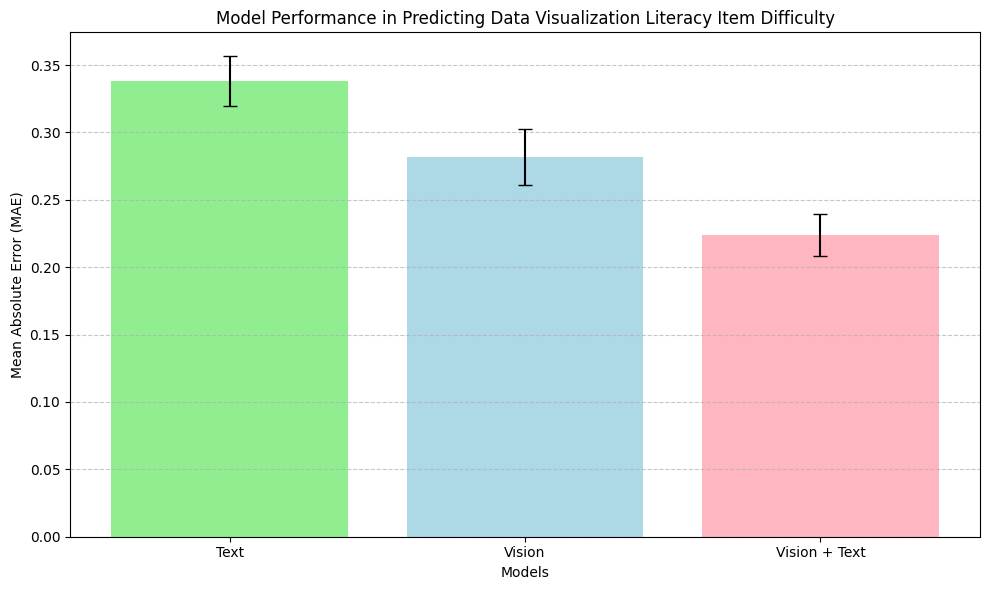} % Replace with the actual filename of your bar chart
  \caption{MAE for each predictive model on the validation subset. Error bars represent the standard error of the mean.}
  \label{fig:mae_bar}
\end{figure}

We applied the three LLM-based models to the subset of 154 items from the validation data with $.png$ images. The MAE for each model was calculated as follows:
\begin{itemize}[noitemsep, nolistsep]
  \item Vision-only Model MAE: 0.2819 
  \item Text-only Model MAE: 0.3382 
  \item Multimodal (Vision + Text) Model MAE: 0.2239 
\end{itemize}
The results in Figure \ref{fig:mae_bar} indicate that the multimodal model achieved the lowest MAE on this validation set, suggesting it provides the most accurate predictions of item difficulty among the three approaches.

\begin{figure}[h]
  \centering
  \includegraphics[width=\linewidth]{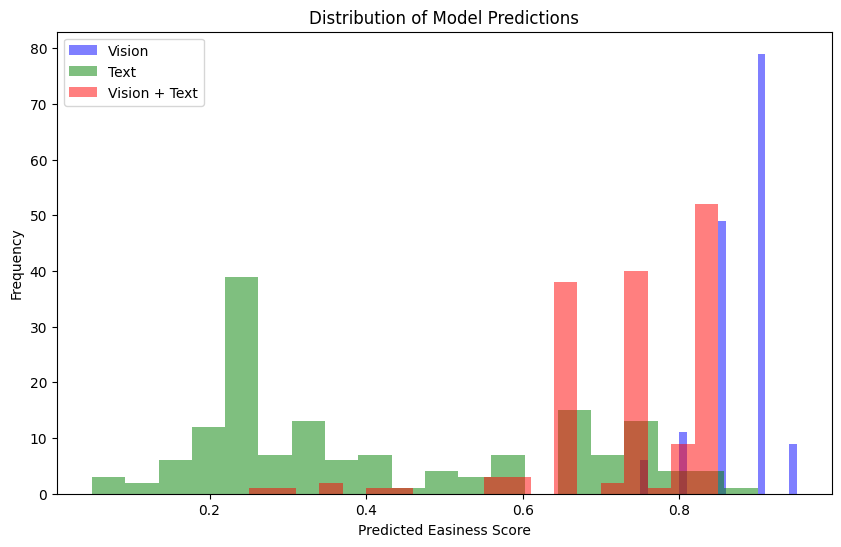} % Replace with the actual filename of your histogram
  \caption{Distribution of predicted easiness scores (proportion correct) for the three models on the validation subset.}
  \label{fig:pred_dist}
\end{figure}

Figure \ref{fig:pred_dist} shows the distribution of predicted easiness scores for each model.
The distribution plot reveals that the vision-only model tends to predict higher easiness scores, with a peak around 0.85-0.9. The Text-only model's predictions are more spread out, with a notable cluster around 0.25. The multimodal model's more centered distribution likely reflects a better balance, capturing the interplay between a clear chart and a complex question, or vice versa.

\subsection{Kaggle Test Set Prediction}
Based on its performance on the validation subset, the multimodal model was selected to make predictions on the held-out test set ($test\_data.csv$). The test set contained 46 items. 6 of these items had $image\_url$s ending in $.svg$, for which direct multimodal processing was not supported by the chosen LLM API at the time of the project. For these 6 items, a default prediction of 0.5 (representing chance performance on a binary item) was assigned. The multimodal model was applied to the remaining 40 items with $.png$ images. The performance of the multimodal model on this held-out test set, as evaluated externally via the competition platform, was reported as a MSE of 0.10805.

\section{Discussion}
Validation results showed that the multimodal approach best predicted item difficulty. This finding supports the intuition that understanding how a question relates to a specific visual representation is key to task performance, rather than relying solely on the appearance of the graphic or the wording of the question in isolation. The subsequent application of the multimodal model to the held-out test set, yielding an MSE of 0.10805, provides an initial measure of its generalization capability on unseen items. 

\subsection{Limitations}
A limitation of this project was the inability to directly process $.svg$ image files with the chosen multimodal LLM API. This necessitated assigning a default prediction value (0.5) to the 6 $.svg$ items in the test set. This fallback strategy does not leverage any information from these items and likely negatively impacts the overall reported MSE for the test set. Future work should address this by implementing SVG-to-PNG conversion or utilizing APIs that support SVG inputs. The reliance on a single, proprietary LLM also presents a limitation, as performance may vary across models and platforms \citep{bommasani2021}. The validation subset size, while sufficient for initial model comparison, could be expanded for more robust evaluation. Finally, the current model provides only a point prediction for easiness; incorporating measures of prediction uncertainty would be valuable for practical application.

\subsection{Implications}
Despite these limitations, the results suggest that multimodal LLMs hold significant potential for automating psychometric analysis in the domain of data visualization literacy \citep{mislevy2003}. An MSE of 0.10805 on an unseen test set indicates that the multimodal model has learned features generalizable beyond the training data. This capability could greatly accelerate the test development process, allowing for automated pre-calibration of item difficulties and potentially informing item design. Understanding the interplay between visual and textual features, as captured by the LLM's analysis, can also provide insights into common sources of difficulty, which can then guide the creation of more effective educational materials and data visualization design guidelines \citep{munzner2014}. Future research should explore more sophisticated methods for handling diverse image formats, investigating alternative LLM architectures or fine-tuning strategies, and comparing these AI-driven approaches against traditional psychometric modeling techniques.

\bibliographystyle{ACM-Reference-Format}

\end{document}